\documentclass[10pt,twocolumn,letterpaper]{article}

\pdfoutput=1 

\usepackage{cvpr}              

\usepackage{graphicx}
\usepackage{amsmath}
\usepackage{amssymb}
\usepackage{booktabs}
\usepackage{subcaption}
\usepackage{tabularx}
\usepackage{bbold}
\usepackage[table,xcdraw]{xcolor}
\usepackage{textcomp}  
\usepackage{scalerel}  
\usepackage{enumitem}
\usepackage[accsupp]{axessibility}  
\usepackage{xparse}

\definecolor{colbest}{rgb}{0.1, 0.6, 0.1}
\definecolor{colworst}{rgb}{0.75, 0, 0}
\definecolor{oursrow}{HTML}{E6F2F4}

\usepackage{pifont}

\NewDocumentCommand\icon{}{\scalerel*{\includegraphics{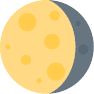}}{X}}
\newcommand{\dataname}{PHASE~\icon~}

\usepackage[pagebackref,breaklinks,colorlinks]{hyperref}

\usepackage[capitalize]{cleveref}
\crefname{section}{Sec.}{Secs.}
\Crefname{section}{Section}{Sections}
\Crefname{table}{Table}{Tables}
\crefname{table}{Tab.}{Tabs.}

\def\cvprPaperID{-} 
\def\confName{CVPR}
\def\confYear{2023}

\begin{document}

\title{Uncurated Image-Text Datasets: Shedding Light on Demographic Bias}

\author{Noa Garcia \hspace{35pt} Yusuke Hirota \hspace{35pt} Yankun Wu \hspace{35pt} Yuta Nakashima\\
{\tt\small \{noagarcia@, y-hirota@is., yankun@is., n-yuta@\}ids.osaka-u.ac.jp}\\
Osaka University
}
\maketitle

\begin{abstract}
The increasing tendency to collect large and uncurated datasets to train vision-and-language models has raised concerns about fair representations. It is known that even small but manually annotated datasets, such as MSCOCO, are affected by societal bias. This problem, far from being solved, may be getting worse with data crawled from the Internet without much control. In addition, the lack of tools to analyze societal bias in big collections of images makes addressing the problem extremely challenging. 

Our first contribution is to annotate part of the Google Conceptual Captions dataset, widely used for training vision-and-language models, with four demographic and two contextual attributes. Our second contribution is to conduct a comprehensive analysis of the annotations, focusing on how different demographic groups are represented. Our last contribution lies in evaluating three prevailing vision-and-language tasks: image captioning, text-image CLIP embeddings, and text-to-image generation, showing that societal bias is a persistent problem in all of them. 

\centerline{\footnotesize \url{https://github.com/noagarcia/phase}}

\end{abstract}

\section{Introduction}
\label{sec:intro}

\begin{figure}[t]
  \centering
  \includegraphics[clip, width=0.9\columnwidth]{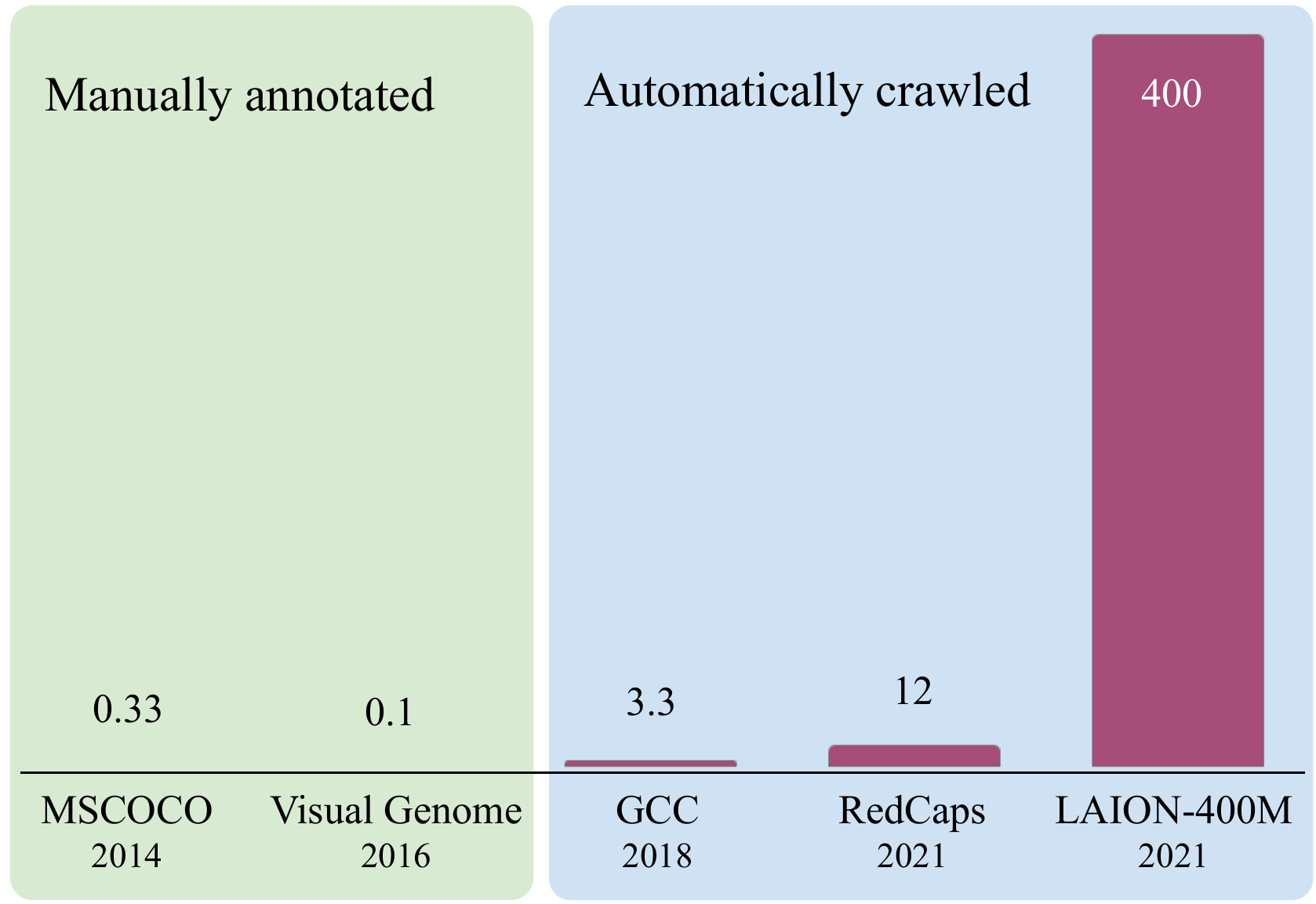}
  \caption{Evolution of paired image-text datasets in terms of number of samples (in million). Datasets scaled up with data automatically crawled from the Internet, reaching the current status in which models are trained with hundreds of millions of samples.}
  \label{fig:vldatasets}
\end{figure}

The training paradigm in vision-and-language models has shifted from manually annotated collections, such as MS-COCO \cite{lin2014microsoft} and Visual Genome \cite{krishna2017visual}, to massive datasets with little-to-none curation automatically crawled from the Internet \cite{sharma2018conceptual, desai2021redcaps, schuhmann2021laion}. Figure \ref{fig:vldatasets} illustrates this tendency by comparing the size of paired image-text datasets over time. Whereas manually annotated datasets, widely used in the last decade, contained a few hundred thousand images each, the latest automatically crawled collections are composed of several million samples. This large amount of data has led to training some  disruptive models in the field, such as CLIP \cite{radford2021learning} trained on 400 million image-text pairs; Imagen \cite{saharia2022photorealistic} trained on 860 million image-text pairs; Flamingo \cite{alayrac2022flamingo} trained on 2.3 billion images and short videos paired with text; DALL-E 2 \cite{ramesh2022hierarchical} trained on 650 million images; or Stable Diffusion \cite{rombach2022high}, trained on 600 million captioned images. Those models have been shown to learn visual and language representations that outperform the previous state-of-the-art on tasks such as zero-shot classification \cite{radford2021learning} or text-to-image generation \cite{ramesh2022hierarchical,rombach2022high}. 

Despite the impressive results on controlled benchmarks, a critical drawback arises: the larger the training set, the less control over the data. With toxic content easily accessible on the Internet, models trained under uncurated collections are more prone to learn  harmful representations of the world, including societal bias, which results in models performing differently for different sociodemographic groups \cite{weidinger2022taxonomy}. 
The risk of obtaining unfair representations is high, as not only do models trained on biased datasets learn to reproduce bias but also amplify it by making predictions more biased than the original data \cite{wang2019balanced,wang2021directional,hirota2022quantifying}. This turns out to be harmful when, far from controlled research environments, models are used in the real-world \cite{buolamwini2018gender}. 

\begin{table*}[t]
\renewcommand{\arraystretch}{1.1}
\setlength{\tabcolsep}{10pt}
\small
\centering
\caption{Image-text datasets with annotations for bias detection.}
\begin{tabularx}{0.94\textwidth}{l l l l r r p{15mm}}
\toprule
Dataset & Image source & Annotation process & Labels & Images & Regions & Attributes\\
\midrule 

Zhao et al. {\protect\NoHyper\cite{zhao2017men}\protect\endNoHyper} & MSCOCO {\protect\NoHyper\cite{lin2014microsoft}\protect\endNoHyper} & automatic (captions) & image & $33,889$ & - & gender \\[0.5cm]

Zhao et al. {\protect\NoHyper\cite{zhao2021captionbias}\protect\endNoHyper} & MSCOCO {\protect\NoHyper\cite{lin2014microsoft}\protect\endNoHyper} (val) & crowd-sourcing & region & $15,762$ & $28,315$ & gender \newline skin-tone\\[0.5cm]

\rowcolor{oursrow}
\dataname & GCC {\protect\NoHyper\cite{sharma2018conceptual}\protect\endNoHyper} & crowd-sourcing & region & $18,889$ & $35,347$  & age \newline gender \newline skin-tone \newline ethnicity \newline emotion \newline activity\\

\bottomrule
\end{tabularx}
\label{tab:datasets}
\end{table*}

Manually annotated datasets \cite{deng2009imagenet,lin2014microsoft} have been shown to be affected by societal bias \cite{zhao2017men,zhao2021captionbias,hirota2022gender,meister2022gender}, but the problem gets worse in automatically crawled datasets \cite{prabhu2020large,birhane2021multimodal}. To overcome societal bias, fairness protocols must be included both in the dataset and in the model development phase. Data analysis \cite{prabhu2020large,wang2020revise,yang2020towards,birhane2021multimodal,meister2022gender,hirota2022gender}, evaluation metrics \cite{wang2021directional,ross2021measuring,hirota2022quantifying}, and mitigation techniques \cite{burns2018women,wang2021gender,berg2022prompt,hirota2023model} are essential tools for developing fairer models, however, they require demographic attributes, such as gender or skin-tone, to be available. These annotations are currently scarce, and only exist for a few datasets and attributes \cite{zhao2017men,zhao2021captionbias}. 

In this paper, we contribute to the analysis, evaluation, and mitigation of bias in vision-and-language tasks by annotating six types of demographic\footnote{Demographic attributes: age, gender, skin-tone, ethnicity.} and contextual\footnote{Contextual attributes: emotion, activity.} attributes in a large dataset: the Google Conceptual Captions (GCC) \cite{sharma2018conceptual}, which was one of the first automatically crawled datasets with 3.3 million image-caption pairs. We name our annotations \dataname (\underline{P}erceived \underline{H}uman \underline{A}nnotations for \underline{S}ocial \underline{E}valuation), and we use them to conduct a comprehensive analysis of the distribution of demographic attributes on the GCC dataset. We complement our findings with experiments on three main vision-and-language tasks: image captioning, text-image embeddings, and text-to-image generation. Overall, we found that a dataset crawled from the internet like GCC presents big unbalances on all the demographic attributes under analysis. Moreover, when compared against the demographic annotations in MSCOCO by Zhao \etal~\cite{zhao2021captionbias}, GCC has bigger representation gaps in gender and skin-tone. As for the downstream tasks, the three of them show evidence of different performance for different demographic groups.

\section{Related work}

\paragraph{Bias in vision-and-language} Vision-and-language are a set of tasks that deal with data in image and text format. This includes image captioning \cite{vinyals2015show}, visual question answering \cite{antol2015vqa}, or visual grounding \cite{plummer2015flickr30k}. In terms of societal bias, Burns \etal \cite{burns2018women} showed that captions in the standard MSCOCO dataset~\cite{chen2015microsoft} were gender imbalanced, and proposed an equalizer to mitigate the problem. Since then, gender bias has been found not only in image captioning \cite{amend2021evaluating,tang2021mitigating,hirota2022quantifying,wang2022measuring} but also in text-to-image search \cite{wang2021gender}, pretrained vision-and-language models \cite{srinivasan-bisk-2022-worst, berg2022prompt}, multimodal embeddings \cite{ross2021measuring}, visual question answering \cite{hirota2022gender}, and multimodal datasets \cite{birhane2021multimodal}. Zhao \etal \cite{zhao2021captionbias} showed that gender is not the only attribute affected by bias; skin-tone also contributes to getting differences in captions. As the problem is far from being solved, tools to study and mitigate models' different demographic representations are essential.

\vspace{-10pt}

\paragraph{Bias detection datasets}
Annotations for studying societal bias in vision-and-language tasks are scarce. Without enough data, it is unfeasible to  analyze and propose solutions to overcome the problem. Previous work \cite{zhao2017men,zhao2021captionbias} annotated samples from the MSCOCO dataset \cite{lin2014microsoft} with the perceived attributes of the people in the images. First, \cite{zhao2017men} automatically assigned a binary gender class to images by using the gender words from the captions, excluding images whose captions contained multiple genders. Alternatively, \cite{zhao2021captionbias} annotated gender and skin-tone via crowd-sourcing. In this case, the annotations were conducted at the person-level, as opposed to the whole image, allowing images with multiple people to have multiple annotations. To increase diversity in images other than MSCOCO and attributes other than gender and skin-tone, we annotate \dataname from GCC with six attributes, four demographic and two contextual. Image-text datasets with annotations for bias detection are summarised in Table \ref{tab:datasets}.
\section{\dataname annotations}
Manually annotating demographic attributes from images poses many challenges, as discussed in detail in \cite{andrews2023ethical}. The attributes perceived by external observers may not correspond with the real attributes of the annotated person. Moreover, the definition of some attributes, such as the ones related to race or ethnicity, is ambiguous and subjective \cite{hanna2020towards}. Even so, demographic annotations are essential to measuring how imbalanced a dataset or a model's output is. We attempt to mitigate those problems by first, clarifying to both annotators and potential users that the annotations do not correspond to real attributes, but perceived ones; and second, mitigating the effects of subjectivity by collecting multiple annotations per sample, using two race-related attributes instead of one, and sharing the attributes of the anonymized annotators for uncovering potential correlations between annotators background and their perception.

\subsection{Image source}

The GCC dataset \cite{sharma2018conceptual} contains about 3.3 million samples from the Internet paired with alt-text captions. Originally, images were filtered to remove pornography while captions were post-processed to transform named-entities into hypernyms, \eg \textit{Harrison Ford} $ \rightarrow$ \textit{actor}. Other than that, no filters were applied to remove toxicity or balance the representations. Due to its large size, GCC has been used for pre-training several vision-and-language models, including VilBERT \cite{lu2019vilbert}, VLBERT \cite{su2019vl}, Unicoder-VL \cite{Li2020UnicoderVLAU}, UNITER \cite{chen2020uniter}, OSCAR \cite{li2020oscar}, or ERNIE-VL \cite{yu2021ernie}. This makes it an ideal testbed for studying how the representation of different demographic attributes affects downstream tasks. 

\subsection{Attributes}
Following \cite{zhao2021captionbias}, we annotate people in images via crowd-sourcing. Details on the annotation process are provided in Section \ref{sec:dataset:process}. For each person, we use Amazon Mechanical Turk\footnote{\url{https://www.mturk.com/}} (AMT) to get four \textit{demographic} and two \textit{contextual} attributes. With up to six attributes per person, our goal is to analyze bias from an intersectional perspective. 

\vspace{-12pt}
\paragraph{Demographic attributes} We denote demographic attributes as characteristics of people that are intrinsic to their being and cannot be easily changed. We annotate four attributes with the following categorization:\footnote{Demographic categorization systems cannot represent all the different identities, and thus, it should only be seen as a rough and non-inclusive approximation to different social groups in order to analyze disparities. }  %
\begin{itemize}[noitemsep,topsep=0pt]
    \item \textbf{Age} with five classes: \textit{Baby} (0-2 years-old), \textit{Child} (3-14 years-old), \textit{Young adult} (15-29 years-old), \textit{Adult} (30-64 years-old), \textit{Senior} (65 years-old or more).
    \item \textbf{Gender} with two classes: \textit{Man} and \textit{Woman}.
    \item \textbf{Skin-tone} with six types, \textit{Type 1} to \textit{Type 6}, according to the Fitzpatrick scale \cite{fitzpatrick1988validity}.
    \item \textbf{Ethnicity} with eight classes from the FairFace dataset \cite{karkkainen2021fairface}: \textit{Black},\textit{ East Asian}, \textit{Indian}, \textit{Latino}, \textit{Middle Eastern}, \textit{Southeast Asian}, \textit{White} plus an extra \textit{Other} class. 
\end{itemize}

\vspace{-12pt}
\paragraph{Contextual attributes} We denote contextual attributes as temporary states captured in an image. 
We annotate two: %
\begin{itemize}[noitemsep,topsep=0pt]
    \item \textbf{Emotion} with five classes: \textit{Happiness}, \textit{Sadness}, \textit{Fear}, \textit{Anger}, \textit{Neutral}.
    \item \textbf{Activity} with ten groups adapted from the ActivityNet taxonomy \cite{caba2015activitynet} to fit static images: \textit{Helping and Caring}, \textit{Eating}, \textit{Household}, \textit{Dance and Music}, \textit{Personal Care}, \textit{Posing}, \textit{Sports}, \textit{Transportation}, \textit{Work}, and \textit{Other}.
\end{itemize}


\noindent
All the attributes also include an \textit{unsure} class. 

\subsection{Annotation process}
\label{sec:dataset:process}
\begin{figure}[t]
  \centering
  \vspace{-5pt}
  \includegraphics[clip, width=0.9\columnwidth]{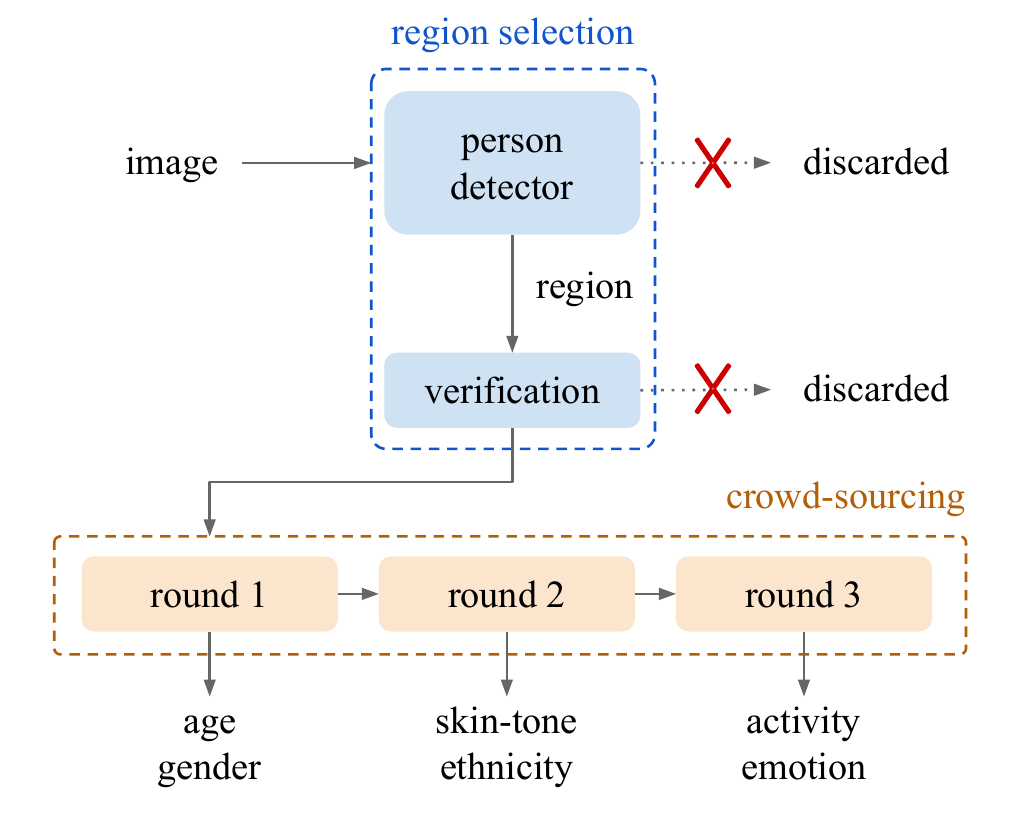}
    \vspace{-5pt}
  \caption{Annotation process. In the region selection part, regions with people are detected and filtered. In the crowd-sourcing part, the selected regions are annotated in three rounds.}
  \label{fig:annotation}
\end{figure}

Due to the large size of the GCC dataset, we annotate all the validation set ($4,614$ images with people) and a random subset of the training set ($14,275$ images with people). The annotation process\footnote{Annotation process approved by the Institutional Review Board (IRB).} consists of two parts, region selection and crowd-sourcing, summarized in Figure \ref{fig:annotation}.

\begin{figure*}[t]
  \centering
  \includegraphics[clip, width=0.98\linewidth]{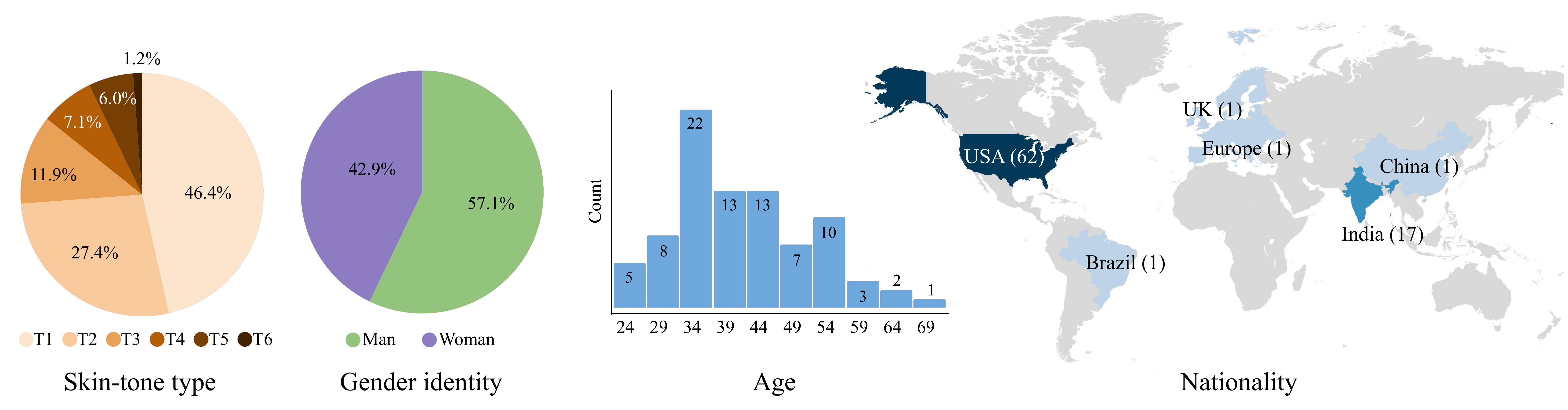}
  \caption{Statistics on workers' self-reported attributes. From left to right: skin-tone type pie chart, gender identity pie chart, age histogram with bin width of $5$ years, and nationality map.}
  \label{fig:workers_demographics}
\end{figure*}

\vspace{-12pt}
\paragraph{Region selection} Unlike MSCOCO, GCC does not have object region annotations. There are machine-generated labels for a subset of the training images. We do not rely on these labels, as they are not associated with bounding boxes and are not available for the validation set. Instead, we conduct a region selection process semi-automatically in two phases. In the first phase, we run the object detector YOLOv5 \cite{glenn_jocher_2022_7002879} to detect regions with objects. For each detected region, YOLOv5 returns an object class, a confidence score, and a bounding box. We keep regions whose object label is \textit{person}, the confidence score is higher than $0.35$, and the pixel area is larger than $5,000$. As the detected regions are relatively noisy, in the second phase, we conduct a manual verification. We discard regions without a person, depicting multiple people, or not photographs.

\vspace{-12pt}
\paragraph{Crowd-sourcing} The selected regions are annotated via AMT crowd-sourcing in three rounds. In the first round, we collect the demographic attributes of age and gender. In the second round, the ones related to race, \ie skin-tone and ethnicity. Lastly, we collect the contextual attributes of emotion and activity. Splitting the annotation process into different rounds allows us to have better control of the quality. For the first two rounds, we expose workers to the cropped regions, to avoid the context being used to predict demographic attributes \cite{bhargava2019exposing}. For the last round, as the context is essential, we show workers the full image with a bounding box around the person of interest. We ask them to select the perceived attributes from a given list. For all rounds, three different workers annotate each region. To ensure quality, we conduct random verifications of the annotated samples.

Workers must fulfill three conditions: 1) to be \textit{Masters}, which is a qualification granted by the platform to workers with high-quality work, 2) to agree to a consent form as per our Institutional Review Board approval, and 3) to conduct a survey for collecting workers' demographics. As the perception of attributes may be affected by the own workers' attributes \cite{davani2022dealing}, we anonymize workers' survey and release it with the rest of the annotations. 
\section{Annotator statistics}
\label{sec:dataset:workers}

\paragraph{Annotator demographics} The workers' survey has the following questions: \textit{age} as an input text box; \textit{gender} as a multiple choice form with options \textit{man}, \textit{woman}, \textit{non-binary}, \textit{others}; \textit{nationality} as an input text box; and \textit{skin-tone type} as a multiple choice form with six options according to the Fitzpatrick scale. Statistics are shown in Figure \ref{fig:workers_demographics}. In total, $84$ workers answered the survey. From these, $9.5\%$ reported to be less then $30$ year-old, $88.1\%$ between $30$ and $64$ year-old, and $2.4\%$ more than 65 year-old. With respect to gender, $42.9\%$ workers reported to be women, and $57.1\%$ men, with no other genders reported. Skin-tone is predominantly light (Types 1, 2, and 3), with only $14.3\%$ skin-tone Types 4, 5, or 6. Most of the workers are from the United States of America ($73.8\%$) followed by India ($20.2\%$). Other nationalities include Brazil, the United Kingdom, Europe, and China, with a worker each. 

\begin{table}[t]
\renewcommand{\arraystretch}{1.1}
\setlength{\tabcolsep}{3pt}
\small
\centering
\caption{Inter-annotator agreement in the training set. 3+ and 2+ indicate the ratio ($\%$) of regions with a consensus of three or more, or two or more workers, respectively. $\kappa$ indicates Fleiss' kappa and Agreement is set according to $\kappa$ as in \cite{viera2005understanding}. Labels indicates the number of classes plus the \textit{unsure} class ($+1$).}
\begin{tabularx}{0.98\columnwidth}{@{} l r r r r l}
\toprule

Attribute & Labels & 3+ & 2+ & $\kappa$ & Agreement\\
\midrule
age & $5 + 1$ & $48.8$ & $97.2$ & $0.44$ & moderate\\
gender & $2 + 1$ & $92.5$ & $99.6$ & $0.89$ & almost perfect\\
skin-tone & $6 + 1$ & $27.1$ & $86.3$ & $0.24$ & fair \\
skin-tone (binary) & $2 + 1$ & $80.0$ & $99.4$ & $0.59$ & moderate \\
ethnicity & $8 + 1$ & $50.9$ & $90.1$ & $0.41$ & moderate \\
emotion & $5 + 1$ & $46.5$ & $94.2$ & $0.37$ &  fair \\
activity & $11 + 1$ & $61.8$ & $95.3$ & $0.65$ & substantial \\

\bottomrule
\end{tabularx}
\label{tab:iaa}
\end{table} 

\vspace{-10pt}
\paragraph{Inter-annotator agreement}
We measure to what extent attributes are equally perceived by different people by computing the inter-annotator agreement. We use two metrics: a consensus ratio and the Fleiss' kappa \cite{Fleiss71}. Consensus ratio, $n+$, indicates the percentage of regions in which $n$ or more workers give the same class, and Fleiss' kappa, $\kappa$, measures whether the annotations agree with a probability above chance. Results are in Table \ref{tab:iaa}. According to $\kappa$, gender has an almost perfect agreement, with $92.5\%$ of the regions with a consensus of three workers. Activity shows a substantial agreement, whereas age and ethnicity have a moderate agreement. Considering that ethnicity is known to be a subjective attribute\cite{hanna2020towards}, the relatively high agreement may be due to workers having a similar background (\eg most workers are from the United States of America). In contrast, skin-tone and emotion have the lowest $\kappa$ ($0.24$ and $0.37$, respectively) but are still well above chance ($\kappa \leq 0$). Note that skin-tone can be affected by image illumination and the annotator's own perception of color, making consecutive skin-tone types (\eg \textit{Type 2} and \textit{Type 3})  difficult to distinguish. Thus, we additionally check agreement in binary skin-tone classification, \ie lighter skin-tone (\textit{Types 1, 2, 3}) and darker skin-tone (\textit{Types 4, 5, 6}). $\kappa$ increases from $0.24$ (fair) to $0.59$ (moderate), so from now on we preferably use binary skin-tone unless otherwise stated.  

\section{\dataname analysis}
\label{sec:analysis}

We annotated the whole validation set and a random portion of the training set. We downloaded $13,501$ and $2,099,769$ validation and training images, respectively; from those, $5,668$ and $498,006$ validation and training images had detected human regions by YOLOv5; from those, all the validation ($5,668$) and a random subset of the training images ($17,147$), were annotated; only $4,614$ and $14,275$ validation and training images passed the manual verification and their human-detected regions were annotated with the six demographic and contextual attributes. Overall, $35,347$ regions were annotated: $8,833$ in the validation set and $26,514$ in the training set.

We publicly share the data in two formats: \textit{raw}, in which each region has up to three annotations, and \textit{region-level}, in which each region is assigned to a class by majority vote. If there is no consensus, the region is labeled as \textit{disagreement}.

\vspace{-15pt}
\paragraph{Attribute analysis}
Full statistics per attribute and class are reported in the supplementary material. Overall, all the attributes are imbalanced, with one or two predominant classes per attribute. For age, the predominant class is \textit{adult}, appearing in $45.6\%$ of the region-level annotations, while for gender, the gap between \textit{man} ($63.6\%$) and \textit{woman} ($35.1\%$) is $28.5$ points. The gap is bigger in skin-tone and ethnicity: skin-tone \textit{Type 2} is annotated in $47.1\%$ of the regions, while \textit{Types 4}, \textit{5}, and \textit{6} together only in $17.5\%$. Similarly, in ethnicity, \textit{White} class is over-represented with $62.5\%$ regions, while the rest of the classes appear from $0.6\%$ to $10\%$ regions each. When conducting an intersectional analysis, big differences arise. For example, the classes \textit{man} and \textit{White} appear together in $13,651$ regions, whereas \textit{woman} and \textit{Black} in only $823$.

As per the contextual attributes, the most represented emotion is \textit{neutral} ($47.1\%$) followed by \textit{happy} ($35.7\%$), whereas negative-associated emotions (\textit{sad}, \textit{fear}, \textit{anger}) are just $2.33\%$ in total. For activity, the most common classes are \textit{posing} ($28.6\%$) and \textit{other} ($20.5\%$).



\begin{table*}[t]
\renewcommand{\arraystretch}{1.2}
\setlength{\tabcolsep}{7pt}
\small
\centering
\caption{Image captioning models in terms of bias metrics on \dataname.}
\begin{tabularx}{0.92\textwidth}{@{} l r r c r r r c r r c r r}
\toprule
 & \multicolumn{2}{c}{Age} & & \multicolumn{3}{c}{Gender} & & \multicolumn{2}{c}{Skin-tone (binary)} & & \multicolumn{2}{c}{Ethnicity} \\ 
 \cline{2-3} 
 \cline{5-7} 
 \cline{9-10} 
 \cline{12-13}
 & LIC$_M$ & LIC & & LIC$_M$ & LIC & Error & & LIC$_M$ & LIC & & LIC$_M$ & LIC\\ 
 \midrule
\textit{Unbiased} & \textit{25.0} & \textit{0.0} && \textit{50.0} & \textit{0.0} & \textit{0.0} && \textit{50.0} & \textit{0.0} && \textit{14.3} & \textit{0.0}\\
OFA \cite{wang2022unifying} & 61.1 $\pm$ 3.7 & 7.9 & & 74.7 $\pm$ 3.7 & 2.0 & 4.3 && 60.6 $\pm$ 4.9 & -2.4 && 22.6 $\pm$ 5.4 & 0.9\\
ClipCap \cite{mokady2021clipcap} & 57.4 $\pm$ 3.7 & 6.2 && 76.7 $\pm$ 1.8 & 2.7 & 7.9 && 68.0 $\pm$ 6.5 & 6.2 && 24.8 $\pm$ 7.5 & 2.9\\

\bottomrule
\end{tabularx}
\label{tab:captioning_hard}
\end{table*} 

\begin{figure}[t]
  \centering
  \includegraphics[clip, width=\columnwidth]{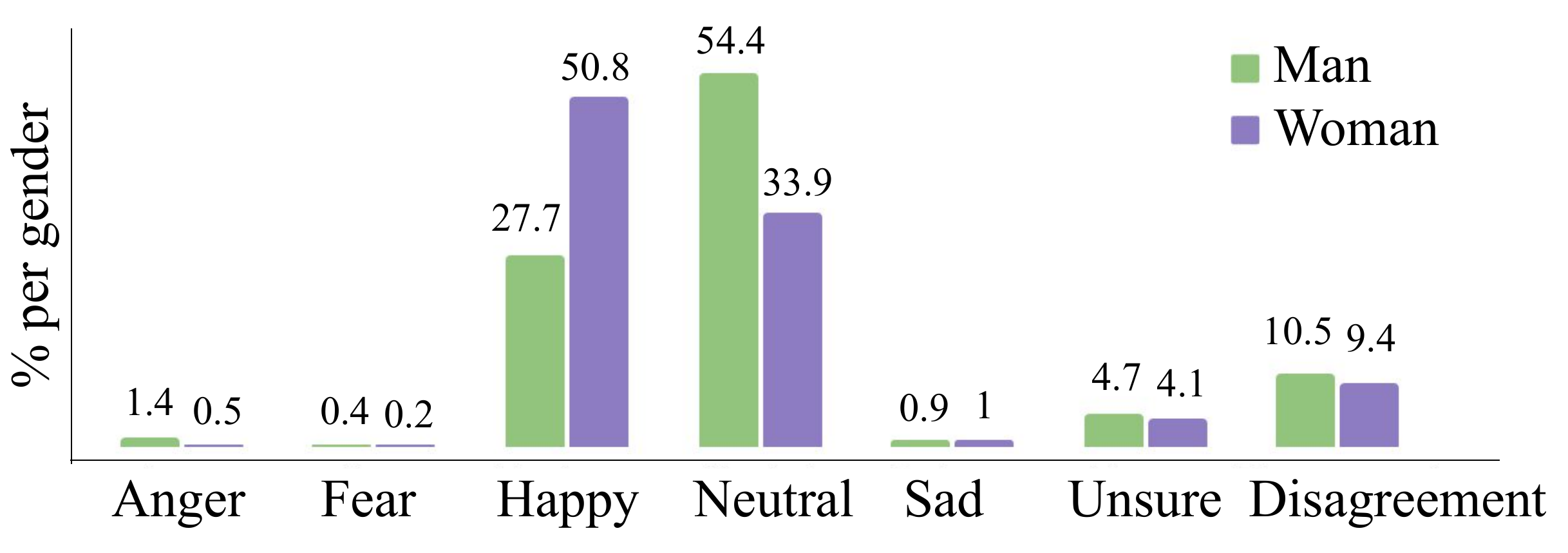}
  \caption{Region-level emotion statistics per gender (percentage).}
  \label{fig:gender-emotion}
\end{figure}

\vspace{-15pt}
\paragraph{Gender and context}
We cross-check gender with contextual attributes. Region-level emotion statistics per gender are shown in Figure \ref{fig:gender-emotion}, where it can be seen that women tend to appear \textit{Happy} whereas men tend to appear \textit{Neutral}. This aligns with the gender stereotyping of emotions, which is a well-documented phenomenon in psychology \cite{plant2000gender}. We also detect disparities in activities per gender,  especially in the classes \textit{posing} and \textit{sports}: in \textit{woman} regions, there are $42.8\%$ \textit{posing} and $5.1\%$ \textit{sports} annotations, while in \textit{man} regions, $21\%$ are \textit{posing} and $26.9\%$ are \textit{sports}.

\vspace{-15pt}
\paragraph{GCC vs MSCOCO} We compare gender and skin-tone annotations in GCC against the annotations in MSCOCO by Zhao \etal \cite{zhao2021captionbias}. In MSCOCO, gender was reported as $47.4\%$ \textit{man} regions and $23.7\%$ \textit{woman} regions, which is a gap of $23.7$ points, smaller than the $28.5$ points gap in GCC.  With respect to skin-tone, the gap in MSCOCO between lighter skin-tone (\textit{Types 1}, \textit{2}, and \textit{3}) and darker skin-tone (\textit{Types 4}, \textit{5}, and \textit{6}) was reported to be $52.9$ (from $61\%$ to $8.1\%$), whereas the gap in GCC is $64$ (from $79.7\%$ to $15.7\%$). This shows that GCC, an automatically crawled dataset, is more unbalanced than MSCOCO, a manually annotated dataset, both in gender and skin-tone attributes. 

\begin{figure}[t]
  \centering
  \includegraphics[clip, width=\columnwidth]{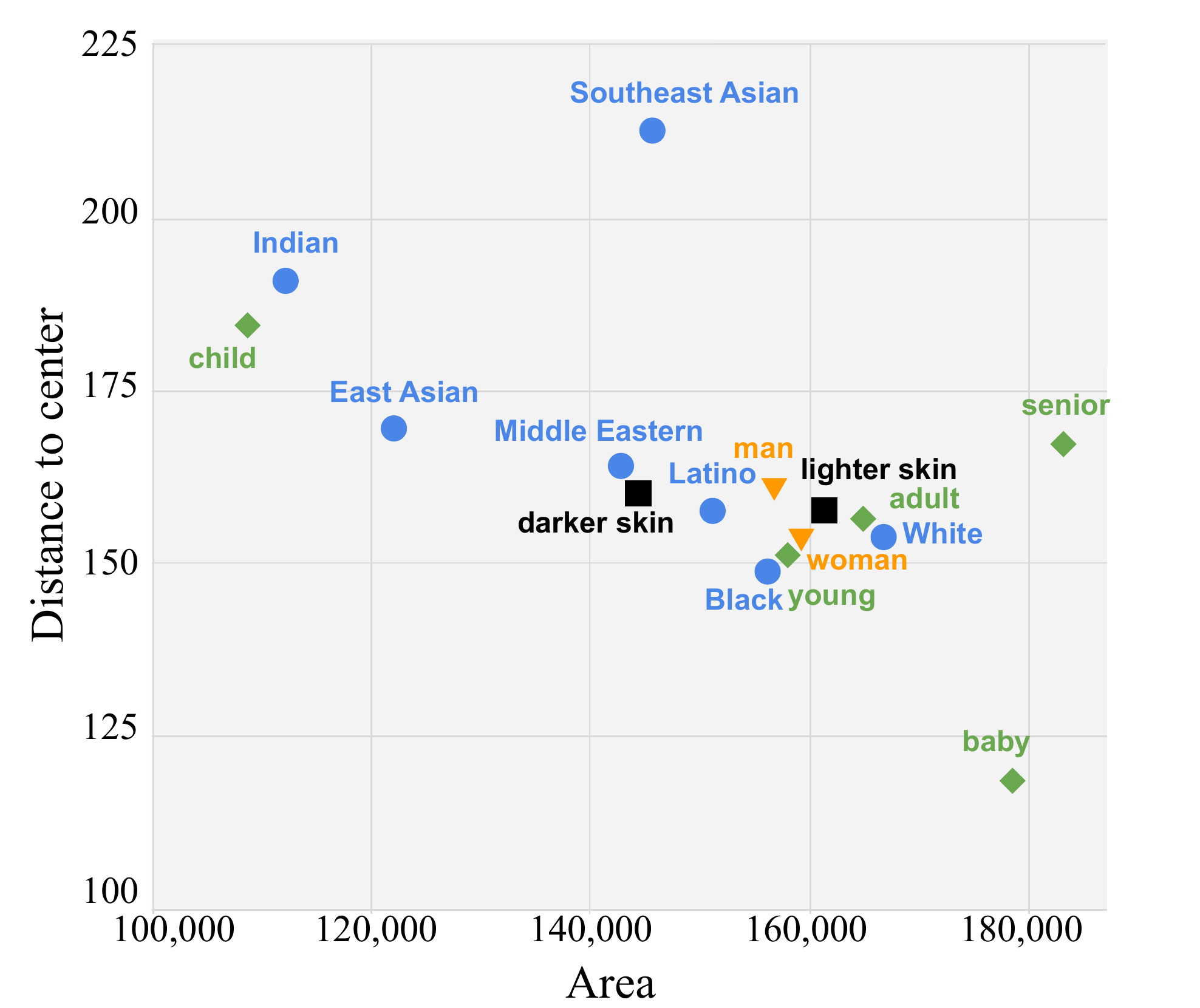}
  \caption{Average area and distance to the image center (in pixels) for each demographic class in the annotated GCC dataset. Colors and shapes indicate classes in the same attribute.}
  \label{fig:area}
\end{figure}

\vspace{-15pt}
\paragraph{Region analysis} We analyze how the regions of the different demographic classes are presented in the GCC dataset. In particular, in Figure \ref{fig:area}, we show an average of the area and the distance to the image center per class. The most noticeable disparities are in the attribute age, with \textit{senior} and \textit{baby} being the larger regions from all the classes and \textit{child} the smaller. The class \textit{baby} also tends to be more centered than people in other ages. In contrast, gender regions do not present big differences in size and position. With respect to skin-tone and ethnicity, the area for \textit{lighter} class is bigger than for \textit{darker} skin-tone, whereas ethnicity presents a large variance, with the \textit{Southeast Asian} class being the most faraway from the center and the \textit{White} class the largest one.
\section{Downstream bias evaluation}
With \dataname annotations we can evaluate societal bias in vision-and-language tasks. We explore image captioning, text-image embeddings, and text-to-image generation.

\vspace{-15pt}
\paragraph{Image annotations} The annotations in \dataname correspond to regions, but the three tasks analyzed in this section use the full image as a whole. To deal with this discrepancy, we transform the region-level annotations into image-level annotations. To that end, we only use images in which all the region-level annotations belong to the same class. For example, an image is labeled as \textit{woman} if all the region-level annotations of gender are \textit{woman}. Images containing different classes for a specific attribute are not used. Note that this approach is not unique and other methods can be used according to the type of evaluation to be conducted. In addition, to deal with the low number of annotations, we merge \textit{baby} and \textit{child} into a single class \textit{baby \& child}. 

\subsection{Image captioning}
\label{sec:captioning}

Image captioning is one of the reference tasks in vision-and-language research. Given an image, a captioning model generates a sentence describing its contents. Multiple image captioning models trained on the MSCOCO dataset have been shown to be biased with respect to gender \cite{burns2018women} and skin-tone \cite{zhao2021captionbias}. In this section, we evaluate two of the latest models trained on GCC, OFA \cite{wang2022unifying} and ClipCap \cite{mokady2021clipcap}, in the four demographic attributes in our annotations. While OFA is a state-of-the-art model based on vision-and-language Transformers \cite{vaswani2017attention}, ClipCap leverages CLIP embeddings \cite{radford2021learning}, which are also analyzed in terms of bias in the next section.

\vspace{-10pt}
\paragraph{Metrics} We evaluate image captioning models on societal bias and bias amplification with LIC$_M$ and LIC metrics \cite{hirota2022quantifying}, respectively. LIC$_M$ corresponds to the accuracy of a caption classifier that predicts the  class of a demographic attribute after masking class-revealing words.%
\footnote{Gender-revealing words: actor, actress, aunt, boy, boyfriend, brother, chairman, chairwoman, cowboy, daughter, dude, emperor, father, female, gentleman, girl, girlfriend, guy, he, her, hers, herself, him, himself, his, husband, lady, male, man, mother, policeman, policewoman, pregnant, prince, princess, queen, king, she, sister, son, uncle, waiter, waitress, wife, woman (and their plurals).}%
\footnote{Age-revealing words: adult, aged, baby, child, elderly, infant, kid, teenager, toddler, young (and their plurals).}
If the caption classifier is more accurate than random chance, it means that captions of people from different classes are semantically different. The classifier is trained $10$ times with different random seeds, and results are reported as average and standard deviation. On the other hand, $\text{LIC} = \text{LIC}_M - \text{LIC}_D$ measures bias amplification by comparing the accuracy of the caption classifier trained on the human captions, LIC$_D$, against the model-generated captions, LIC$_M$. If $\text{LIC}>0$, the generated captions are more biased than the original ones and the model amplifies the bias. 

Additionally, for the gender attribute, we compute Error \cite{burns2018women}, which measures the percentage of captions in which gender is misclassified. Error can only be computed if attributes are explicitly mentioned in the generated captions, which is the case for gender, but not for \eg skin-tone.

\vspace{-10pt}
\paragraph{Results} Results are shown in Table \ref{tab:captioning_hard}. Both OFA and ClipCap have a LIC$_M$ score well above the unbiased case for the four attributes. When trained on GCC, except OFA in skin-tone, both models amplify bias with respect to the original dataset. This highlights the urgency of incorporating bias mitigation techniques, such as the model-agnostic method in \cite{hirota2023model}. Although age is an attribute that is not often analyzed, results show that it is the one with the highest bias, both on LIC$_M$ and LIC metrics. This highlights the urgency to consider age in representation fairness. Results on gender and skin-tone also reveal important biases in the models' outputs, while the large standard deviation in ethnicity, which may be due to the higher number of classes ($7$) and the smaller number of samples per class ($64$), makes it difficult to extract reliable conclusions.

\subsection{Text-image CLIP embeddings}

\begin{table}[t]
\renewcommand{\arraystretch}{1.2}
\setlength{\tabcolsep}{5pt}
\small
\centering
\caption{CLIP embeddings evaluation on \dataname validation set.}
\begin{tabularx}{0.99\columnwidth}{@{} l l r r r r }
\toprule
Attribute & & Samples & R@1 & R@5 & R@10 \\
\midrule
age & baby \& child & $350$ & $44.0$ & $65.4$ & $74.0$ \\
    & young & $1,349$ & $30.4$ & $51.3$ & $60.9$ \\
    & adult & $1,509$ & $27.3$ & $46.7$ & $55.9$ \\
    & senior & $128$ & $44.5$ & $64.1$ & $71.1$ \\
\midrule

gender    & man & $1,950$ & $32.0$ & $53.2$ & $63.1$ \\
  & woman & $1,617$ & $30.6$ & $49.8$ & $59.1$ \\

\midrule

skin-tone  & lighter & $3,166$ & $30.2$ & $50.6$ & $59.9$ \\
    & darker & $318$ & $31.1$ & $54.1$ & $62.3$ \\
\midrule    

ethnicity  & Black & $194$ & $29.4$ & $51.5$ & $58.8$ \\
    & East Asian & $58$ & $34.8$ & $56.9$ & $63.8$ \\
    & Indian & $90$ & $34.4$ & $61.1$ & $68.9$ \\
    & Latino & $28$ & $21.4$ & $39.3$ & $50.0$ \\
    & Middle Eastern & $16$ & $31.3$ & $62.5$ & $75.0$ \\
    & Southeast Asian & $16$ & $31.3$ & $37.5$ & $56.3$ \\
    & White & $2,231$ & $30.6$ & $50.6$ & $59.5$ \\

\bottomrule
\end{tabularx}
\label{tab:clip}
\end{table}

Next, we evaluate the performance of pre-trained text-image CLIP embeddings \cite{radford2021learning} for the different demographic attributes. CLIP is a dual architecture with a text encoder and an image encoder that learns image and text embeddings by predicting matching pairs. Due to the large amount of data used for training ($400$ million image-text pairs), the two encoders learn the correspondences of high-level semantics in the language and the visual modalities. The goal of our evaluation is to check whether the demographic attributes of people in the images have an impact on the accuracy of the embeddings. To conduct this evaluation, we extract image and text embeddings with the pre-trained CLIP encoders. For each of the $4,614$ images in the validation set, we rank the validation captions according to the cosine similarity between their embeddings and analyze the accuracy of the ranked list. 

\begin{figure*}[t]
  \centering
  \includegraphics[clip, width=0.98\linewidth]{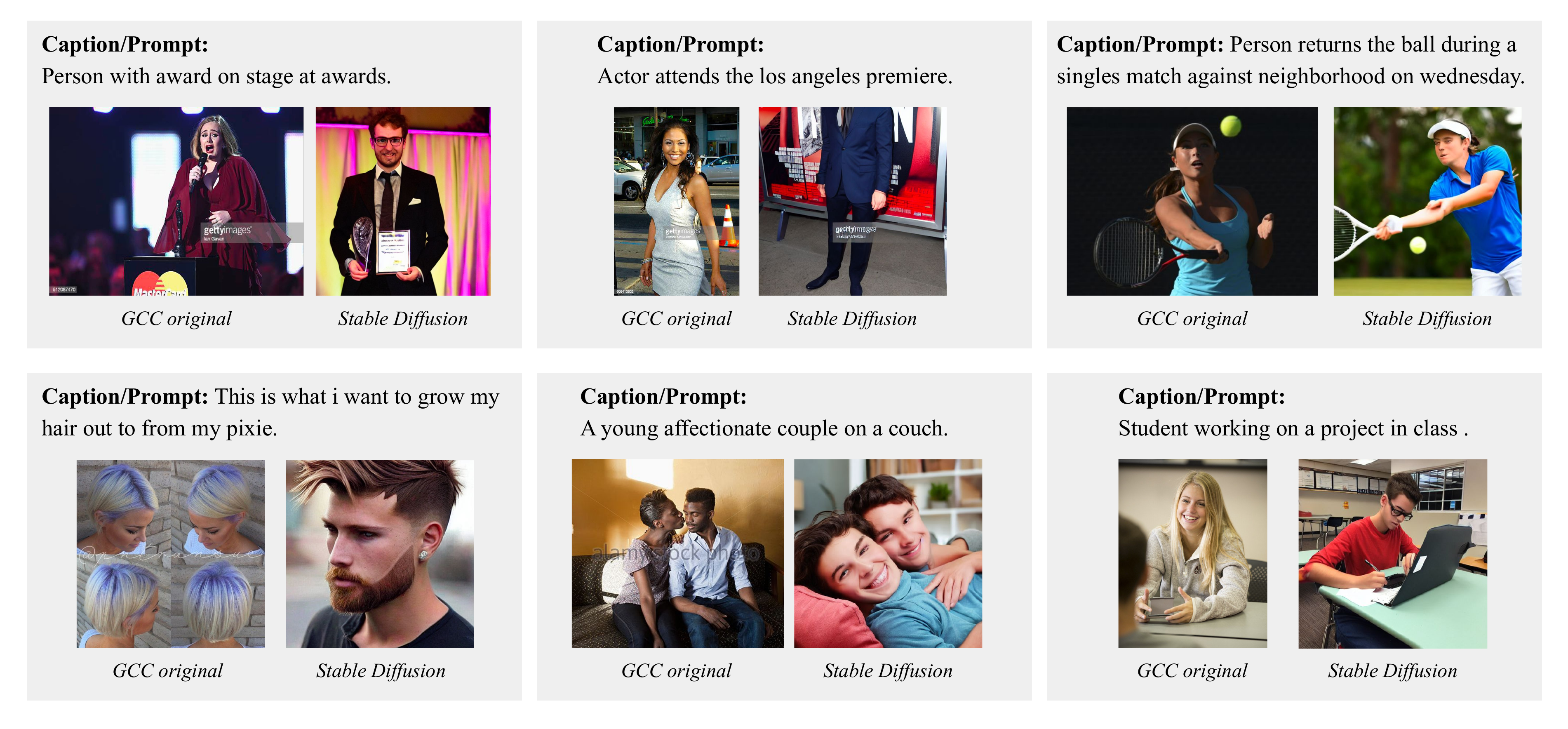}
  \caption{Examples of Stable Diffusion \cite{rombach2022high} generated images with the caption used as a prompt, together with the original GCC image. Stable Diffusion tends to generate images of White men for captions with neutral language (\eg person, student, couple, \etc).}
  \vspace{-10pt}
  \label{fig:diffusion}
\end{figure*}

\vspace{-10pt}
\paragraph{Metrics}  We evaluate accuracy of image-text embeddings as recall at $k$ (R@$k$) with $k = \{1, 5, 10\}$.  R@$k$ indicates the percentage of images whose matching caption is ranked within the top-$k$ positions. We compare the difference in R@$k$ for classes in the same attribute to check whether CLIP embeddings perform differently. In the ideal scenario of unbiased representations, R@$k$ performance in different class attributes should be alike. 

\vspace{-10pt}
\paragraph{Results} Results are shown in Table \ref{tab:clip}. Within each of the four demographic attributes, noticeable differences in performance can be observed. Note that the number of samples in each class is different, which could influence the results. In the supplementary material, we report an evaluation when using the same number of samples per class, in which we verify that the conclusions are not affected. We summarize the main findings as follows:

\begin{itemize}[noitemsep,topsep=0pt,wide]
    \item For the age attribute, \textit{baby \& child} and \textit{senior} have the best R@$k$, while \textit{young} and \textit{adult} fall well behind with a difference of up to $18.1$ in R@10. The big differences in the age classes are consistent with the results of the region analysis (Section \ref{sec:analysis}) and the image captioning (Section \ref{sec:captioning}), being the attribute with the highest class variance.

    \item In gender, \textit{man} samples perform consistently better than \textit{woman} samples.
    
    \item For skin-tone, \textit{darker} has higher R@$k$ than \textit{lighter}. This may be explained by the language bias, a documented phenomenon in which skin-tone descriptions are usually omitted for \textit{lighter} but not for \textit{darker} skin-tones \cite{van2016stereotyping,otterbacher2019we,zhao2021captionbias}. 
    
    \item Ethnicity is not consistent. For R@1, the highest classes are \textit{Eastern Asian} and \textit{Indian}, but for R@5 and R@10 is \textit{Middle Eastern}. The lowest classes are \textit{Latino}, \textit{Southeast Asian}, and \textit{Black}. The lack of consistency, together with the low inter-annotator agreement (Section \ref{sec:dataset:workers}), shows that ethnicity annotations are highly subjective. 

    \item More samples do not ensure a better recall, which means that the number of samples is not the (only) source for difference in performance. For example, in age, although \textit{adult} and \textit{young} classes are the most common, their performance is the worst. The same happens in skin-tone; despite the predominance of \textit{lighter}, R@$k$ is higher for \textit{darker} samples. In contrast, in gender, \textit{man} outperforms \textit{woman} both in the number of samples and in R@$k$.
\end{itemize}

\subsection{Text-to-image generation}
Lastly, we analyze the demographic representation on Stable Diffusion~\cite{rombach2022high}, one of the latest text-to-image generation models. Text-to-image generation, which can be seen as the reverse operation of image captioning, consists on creating an image from a text sentence, also known as \textit{prompt}. In particular, Stable Diffusion relies on pre-trained CLIP embeddings and Diffusion Models \cite{ho2020denoising} to generate an image in the latent space whose embedding is close to the input prompt embedding. In our evaluation, we use the $4,614$ captions in the validation set as prompts to generate an image per caption. We use the demographic annotations of the original images associated with the captions to study Stable Diffusion representations. 

\vspace{-10pt}
\paragraph{Metrics} The official code for Stable Diffusion v1.4\footnote{\url{https://github.com/CompVis/stable-diffusion}} includes a  Safety Checker module that raises a flag when the generated images are considered to be NSFW.\footnote{Not safe for work, a tag commonly used for pornographic, violent, or otherwise inappropriate content.} The module is pre-trained and used off-the-shelf by the community. We check whether there are patterns in the output of the Safety Checker according to the demographic attributes of the input caption. Additionally, we compare the demographics of the generated images against the demographics of the original images associated with the captions.

\vspace{-10pt}
\paragraph{Results} Out of $4,614$ generated images, $36$ are flagged as unsafe by the Safety Checker module. From these, we do not find prominent differences between the distribution of classes in the original images and the unsafe images for age, skin-tone, and ethnicity attributes. We do find, however, the distribution of gender unusual: despite \textit{woman} only being $35.04\%$ validation images, it raises $51.61\%$ unsafe images. This indicates that gender, especially \textit{woman}, has an important contribution to the Safety Checker. With the current experiments, we cannot clarify whether this is due to Stable Diffusion being more prone to generate NSFW images from prompts from women images, or due to the Safety Checker being more prone to detect NSFW in woman-generated images. A table with the results can be found in the supplementary material. Additionally, in Figure \ref{fig:diffusion}, we show some examples of Stable Diffusion generated images and compare them against the original images in GCC. We observe that when the prompt refers to people in a neutral language (\eg. person), the generated images tend to represent \textit{White men}. To verify this observation, we manually annotate 100 generated images. For neutral language prompts, we observe $47.4\%$ men \textit{vs.} $35.5\%$ women; and $54.0\%$ lighter \textit{vs.} $26.6\%$ darker skin-tone. These results are consistent with concurrent work on text-to-image generation bias \cite{bianchi2022easily}.
\section{Limitations}
Although we argued that demographic annotations are necessary to address societal bias in vision-and-language models, we acknowledge that they pose some risks.

\vspace{-15pt}
\paragraph{Perceived attributes} The annotations are conducted by external observers, meaning they reflect perceived attributes. Perceived attributes may not correspond to the real person's attributes. As per dataset construction, GCC in this case but computer vision datasets in general, it is not possible to ask people depicted in the images about their self-perceived attributes. Annotations should not be considered real, objective, or trustworthy labels, but an approximation of how observers classify people in images.

\vspace{-15pt}
\paragraph{Subjectivity} The annotations are subjective and not universal. Many demographic attributes, especially the ones related to race, ethnicity, or emotion, have different classification systems according to different contexts and cultures.

\vspace{-15pt}
\paragraph{Malicious uses} The intended use of the annotations is for research on societal bias and fairness. Although we cannot control who, when, and how will use the annotations once they are publicly available, the use for malicious applications is strictly prohibited. 

\section{Conclusion}
We studied social bias in large and uncurated vision-and-language datasets. Specifically, we annotated part of the GCC dataset with four demographic and two contextual attributes. With annotations on age, gender, skin-tone, ethnicity, emotion, and activity, we conducted a comprehensive analysis of the representation diversity of the dataset. We found all six attributes to be highly unbalanced. When compared against manually annotated datasets such as MSCOCO, GCC presented bigger gaps in gender and skin-tone, with an overrepresentation of \textit{man} and \textit{lighter} skin-tones. Additionally, we evaluated three downstream tasks: image captioning, image-text CLIP embeddings, and text-to-image generation. In all the tasks, we found differences in performance for images in different demographic classes, highlighting the need for resources and solutions.\footnote{This work is partly supported by JST CREST Grant No. JPMJCR20D3, JST FOREST Grant No. JPMJFR216O, JSPS KAKENHI No. JP22K12091, and Grant-in-Aid for Scientific Research (A).}

{\small
\bibliographystyle{ieee_fullname}
\bibliography{egbib}
}

\newpage
\appendix

\section{\dataname statistics}
\label{sec:sup:stats}

Statistics about the number of annotations per attribute and class are reported in Table \ref{tab:class_stats}. For each class, we show the \textit{number of annotations} as the raw annotations for the given class, the \textit{number of regions} as the region-level annotations reached after annotator majority voting, and the \textit{number of images} as the images with at least one region-level annotation with the class of interest.

\section{YOLOv5 bias evaluation}
\label{sec:sup:yolo}
We evaluate YOLOv5 in terms of skin-tone bias to check whether the use of this model can be a contributing factor to the representation discrepancies in \dataname annotations. We detect people in MSCOCO and compare accuracy per skin-tone using \cite{zhao2021captionbias} annotations. Results are reported as follows: \textit{darker} skin-tone recall is $0.49$, and \textit{lighter} skin-tone recall is $0.55$. This shows that there is, indeed, a difference in performance according to skin-tone. However, we believe that it is not as big as to justify the representation gap that was found in the GCC dataset and the conclusions of our analysis still stand.

\begin{table}[t]
\renewcommand{\arraystretch}{1.1}
\setlength{\tabcolsep}{5pt}
\small
\centering
\caption{CLIP embeddings evaluation on \dataname validation set when detected person bounding boxes are occluded (black).}
\begin{tabularx}{0.99\columnwidth}{@{} l l r r r r }
\toprule
Attribute & & Samples & R@1 & R@5 & R@10 \\
\midrule
age & baby \& child & $350$ & $12.9$ & $23.4$ & $29.1$ \\
    & young & $1,349$ & $8.6$ & $16.7$ & $21.9$ \\
    & adult & $1,509$ & $9.4$ & $19.9$ & $25.6$ \\
    & senior & $128$ & $11.7$ & $28.9$ & $34.4$ \\
\midrule

gender    & man & $1,950$ & $12.0$ & $24.3$ & $30.6$ \\
  & woman & $1,617$ & $8.1$ & $15.3$ & $20.0$ \\

\midrule

skin-tone  & lighter & $3,166$ & $8.5$ & $17.5$ & $22.9$ \\
    & darker & $318$ & $11.6$ & $25.2$ & $29.6$ \\
\midrule    

ethnicity  & Black & $194$ & $9.3$ & $20.1$ & $23.2$ \\
    & East Asian & $58$ & $8.6$ & $31.0$ & $36.2$ \\
    & Indian & $90$ & $11.1$ & $28.9$ & $35.6$ \\
    & Latino & $28$ & $7.1$ & $14.3$ & $17.9$ \\
    & Middle Eastern & $16$ & $12.5$ & $12.5$ & $25.0$ \\
    & Southeast Asian & $16$ & $12.5$ & $12.5$ & $12.5$ \\
    & White & $2,231$ & $8.7$ & $17.4$ & $23.0$ \\

\bottomrule
\end{tabularx}
\label{tab:clip_occlusions}
\end{table}

\section{CLIP evaluation results}
\label{sec:sup:clipresults}
 We report the results of the CLIP evaluation when balancing the number of samples per class and attribute. Table \ref{tab:clip_all} compares CLIP performance in R@$k$ for $k=1, 5, 10$ when using all the samples in the validation set (Unbalanced), and when using the same number of samples per class and attribute (Balanced). For the balanced results, we use the number of samples in the smallest class in each attribute. Results are reported as mean and standard deviation over $100$ runs with different random samples per class. 

\begin{table}[t]
\renewcommand{\arraystretch}{1.1}
\small
\centering
\caption{Classes in the validation set and classes labeled as unsafe by Stable Diffusion's Safety Checker.}
\begin{tabularx}{0.94\columnwidth}{@{} l l r r }
\toprule
\multicolumn{2}{l}{Attribute} & Validation set ($\%$) & Unsafe label ($\%$) \\
\midrule

\rowcolor{oursrow}
\multicolumn{2}{l}{age} & & \\
    & baby & $0.89$ & $3.23$  \\
    & child & $6.70$ & $12.90$ \\
    & young & $29.24$ & $32.26$ \\
    & adult & $32.70$ & $32.26$ \\
    & senior & $2.77$ & $3.23$ \\
    & \textit{unsure} & $\textit{1.11}$ & $\textit{0.00}$ \\
    & \textit{multiple} & $\textit{26.59}$ & $\textit{16.13}$\\

\rowcolor{oursrow}
\multicolumn{2}{l}{gender} & & \\
    & man & $42.26$ & $35.48$ \\
  & woman & $35.04$ & $51.61$ \\
    & \textit{unsure} & $\textit{0.98}$ & $0.00$ \\
    & \textit{multiple} & $\textit{21.72}$ & $12.90$ \\

\rowcolor{oursrow}
\multicolumn{2}{l}{skin-tone (binary)} & & \\
  & lighter & $68.62$ & $80.65$ \\
    & darker & $6.89$ & $3.23$ \\
    & \textit{unsure} & $\textit{5.66}$ & $\textit{3.23}$ \\
    & \textit{multiple} & $\textit{18.83}$ & $\textit{12.90}$ \\

\rowcolor{oursrow}
\multicolumn{2}{l}{ethnicity} & & \\
  & Black & $4.20$ & $3.23$ \\
    & East Asian & $1.26$ & $0.00$\\
    & Indian & $1.95$ & $3.23$ \\
    & Latino & $0.61$ & $0.00$ \\
    & Middle Eastern & $0.35$ & $0.00$\\
    & Southeast Asian & $0.35$ & $0.00$ \\
    & White & $48.35$ & $64.52$ \\
    & \textit{unsure} & $\textit{5.52}$ & $\textit{3.23}$ \\
    & \textit{multiple} & $\textit{37.41}$ & $\textit{25.81}$ \\
\bottomrule
\end{tabularx}
\label{tab:safety}
\end{table}

\paragraph{CLIP with occlusions} To better understand what about the image leads to differing performances, we occlude the detected bounding boxes and repeat the evaluation process. We find that masking people bounding boxes makes CLIP's R@1 drop from about 30\% to 8\% for all the attributes, which means that relevant information is contained in person regions. Moreover, as shown in Table \ref{tab:clip_occlusions}, the conclusions are maintained, \eg recall for \textit{man} is higher than for \textit{woman}, which suggests that part of the bias is from the language.

\section{Stable Diffusion results}
\label{sec:sup:stableresults}
Table \ref{tab:safety} reports the statistics of Stable Diffusion's Safety Checker per attribute and class. We compare the percentage of samples per class in the image annotations, with the percentage of samples per class labeled as unsafe by the Safety Checker. It stands out that the class \textit{woman} raises $51.61\%$ of the unsafe labels whereas only accounts for $35.04\%$ of the original images. \textit{Lighter} skin-tone and \textit{White} ethnicity also show big increases in the percentage of samples raised as unsafe, but differently from the \textit{woman} class, both of them are the predominant class in their respective attributes.

\begin{table*}[t]
\vspace{-10pt}
\renewcommand{\arraystretch}{1.3}
\setlength{\tabcolsep}{7pt}
\small
\centering
\caption{Statistics of annotations in \dataname per attribute and class. \textit{Annotations} reports the raw number of annotations per class. \textit{Regions} is the total number of region-level annotations after majority voting. \textit{Images} accounts for the number of images with at least one region-level annotation with the class. Due to the inter-annotator agreement results, skin-tone region-level annotations are conducted for binary skin-tone only. For each attribute, the most common class is highlighted in \textbf{bold} and the unsure class  in \textit{italics}.}
\begin{tabularx}{0.98\textwidth}{@{}p{5mm} l r r r c p{5mm} l r r r}
\toprule
\multicolumn{2}{l}{Attribute} & Annotations & Regions & Images & & 
\multicolumn{2}{l}{Attribute} & Annotations & Regions & Images\\
\midrule

\multicolumn{2}{l}{\cellcolor{oursrow} age} & \cellcolor{oursrow} $106,041$ & \cellcolor{oursrow} $35,347$ & \cellcolor{oursrow} $18,889$ & & \multicolumn{2}{l}{\cellcolor{oursrow} skin-tone type} & \cellcolor{oursrow} $105,801$ & \cellcolor{oursrow} - & \cellcolor{oursrow} -\\
& baby & $955$ & $306$ & $259$ & & & type 1 & $15,388$ & - & -\\
& child & $7,829$ & $2,578$ & $1,569$ & & & \textbf{type 2} & $\textbf{49,821}$ & - & -\\
& young adult & $40,398$ & $13,313$ & $8,841$ & & & type 3 & $18,083$ & - & -\\
& \textbf{adult} & $\textbf{48,604}$ & $\textbf{16,117}$ & $\textbf{10,631}$ & & & type 4 & $8,219$ & - & -\\
& senior & $4,632$ & $1,375$ & $1,152$ & & & type 5 & $5,771$ & - & -\\
& \textit{unsure} & $\textit{3,623}$ & $\textit{653}$ & $\textit{525}$ & & & type 6 & $4,570$ & - & -\\
\multicolumn{2}{l}{\cellcolor{oursrow} gender} & \cellcolor{oursrow} $106,041$ & \cellcolor{oursrow} $35,347$ & \cellcolor{oursrow} $18,889$ & & & \textit{unsure} & $\textit{3,949}$ & $\textit{-}$ & $\textit{-}$\\
& \textbf{man} & $\textbf{67,122}$ & $\textbf{22,491}$ & $\textbf{13,511}$ & & \multicolumn{2}{l}{\cellcolor{oursrow} skin-tone (binary)} & \cellcolor{oursrow} - & \cellcolor{oursrow} $35,347$ & \cellcolor{oursrow} $18,889$\\
& woman & $36,936$ & $12,406$ & $8,329$ & & & \textbf{lighter} & - & $\textbf{28,187}$ & $\textbf{16,245}$\\
& \textit{unsure} & $\textit{1,983}$ & $\textit{285}$ & $\textit{241}$ & & & darker & - & $5,572$ & $3,838$\\
\multicolumn{2}{l}{\cellcolor{oursrow}ethnicity} & \cellcolor{oursrow} $105,801$ & \cellcolor{oursrow} \cellcolor{oursrow} $35,347$ & \cellcolor{oursrow} $18,889$ & & & \textit{unsure} & $\textit{-}$ & $\textit{922}$ & $\textit{730}$\\
& Black & $11,314$ & $3,664$ & $2,657$ & & \multicolumn{2}{l}{\cellcolor{oursrow}activity} & \cellcolor{oursrow} $97,021$ & \cellcolor{oursrow} $35,347$ & \cellcolor{oursrow} $18,889$\\
& East Asian & $3,957$ & $953$ & $707$ & & & caring & $786$ & $127$ & $119$\\
& Indian & $4,434$ & $980$ & $616$ & & & music & $14,706$ & $5,012$ & $3,218$\\
& Latino & $8,826$ & $1,309$ & $1,168$ & & & eating & $1,012$ & $278$ & $206$\\
& Middle Eastern & $5,513$ & $373$ & $349$ & & & household & $180$ & $19$ & $18$\\
& Southeast Asian & $2,706$ & $211$ & $174$ & & & personal & $291$ & $39$ & $33$\\
& \textbf{White} & $\textbf{63,253}$ & $\textbf{22,098}$ & $\textbf{13,698}$ & & & \textbf{posing} & $\textbf{30,409}$ & $\textbf{10,121}$ & $\textbf{6,619}$\\
& \textit{unsure} & $\textit{5,578}$ & $\textit{1,289}$ & $\textit{1,021}$ & & & sports & $19,933$ & $6,725$ & $3,807$\\
\multicolumn{2}{l}{\cellcolor{oursrow}emotion} & \cellcolor{oursrow} $100,248$ & \cellcolor{oursrow} $35,347$ & \cellcolor{oursrow} $18,889$ & & & transportation & $811$ & $224$ & $181$\\
& happy & $41,059$ & $12,603$ & $8,215$ & & & work & $5,043$ & $1,433$ & $891$\\
& sad & $2,221$ & $331$ & $308$ & & & sports & $19,933$ & $6,725$ & $3,807$\\
& fear & $1,205$ & $117$ & $114$ & & & other & $22,770$ & $7,249$ & $4,247$\\
& anger & $2,391$ & $377$ & $346$ & & & \textit{unsure} & $\textit{1,080}$ & $\textit{164}$ & $\textit{149}$\\
& \textbf{neutral} & $\textbf{47,367}$ & $\textbf{16,646}$ & $\textbf{10,473}$ & & \\
& \textit{unsure} & $\textit{6,005}$ & $\textit{1,663}$ & $\textit{1,224}$ & & \\

\bottomrule
\end{tabularx}
\label{tab:class_stats}
\end{table*}
\begin{table*}[t]
\renewcommand{\arraystretch}{1.3}
\setlength{\tabcolsep}{6pt}
\small
\centering
\caption{CLIP evaluation on \dataname validation set. Results are reported as R@$k$ for $k = 1, 5, 10$. \textit{Unbalanced} denotes when all the samples are used, resulting in highly unbalanced classes. \textit{Balanced} denotes when using the same number of samples per class and attribute.}
\begin{tabularx}{0.94\textwidth}{@{} l l c r r r r c r r r r}
\toprule
& & & \multicolumn{4}{c}{Unbalanced} & & \multicolumn{4}{c}{Balanced} \\
\cline{4-7} \cline{9-12}
Attribute & Class & & Size & R@1 & R@5 & R@10 & & Size & R@1 & R@5 & R@10 \\
\midrule
age & baby \& child &  & $350$ & $44.0$ & $65.4$ & $74.0$ & & $128$ & $44.0 \pm 3.5$ & $65.3 \pm 3.4$ & $73.9 \pm 3.1$\\
    & young &  & $1,349$ & $30.4$ & $51.3$ & $60.9$ & & $128$ & $29.8 \pm 3.9$ & $51.0 \pm 4.5$ & $60.8 \pm 4.4$\\
    & adult & & $1,509$ & $27.3$ & $46.7$ & $55.9$ & & $128$ & $27.5 \pm 3.8$ & $46.5 \pm 4.2$ & $55.4 \pm 4.1$\\
    & senior & & $128$ & $44.5$ & $64.1$ & $71.1$ & & $128$ & $44.5 \pm 0.0$ & $64.1 \pm 0.0$ & $71.1 \pm 0.0$\\
\midrule

gender    & man & & $1,950$ & $32.0$ & $53.2$ & $63.1$ & & $1,617$ & $32.1 \pm 0.4$ & $53.2 \pm 0.5$ & $63.1 \pm 0.4$  \\
  & woman & & $1,617$ & $30.6$ & $49.8$ & $59.1$  &  & $1,617$ & $30.6 \pm 0.0$ & $49.8 \pm 0.0$ & $59.1 \pm 0.0$ \\

\midrule

skin-tone  & lighter & & $3,166$ & $30.2$ & $50.6$ & $59.9$ & & $318$ & $30.1 \pm 2.6$ & $50.4 \pm 2.8$ & $60.1 \pm 2.8$ \\
    & darker & & $318$ & $31.1$ & $54.1$ & $62.3$ & & $318$ & $31.1 \pm 0.0$ & $54.1 \pm 0.0$ & $62.3 \pm 0.0$ \\
\midrule    

ethnicity  & Black & & $194$ & $29.4$ & $51.5$ & $58.8$ & & $16$ & $29.1 \pm 11.5$ & $49.9 \pm 11.2$ & $57.7 \pm 11.4$ \\
    & East Asian & & $58$ & $34.8$ & $56.9$ & $63.8$ & & $16$ & $33.5 \pm 10.5$ & $56.4 \pm 10.5$ & $64.2 \pm 10.7$ \\
    & Indian & & $90$ & $34.4$ & $61.1$ & $68.9$ & & $16$ & $34.4 \pm 12.3$ & $61.9 \pm 11.8$ & $69.5 \pm 11.4$ \\
    & Latino & & $28$ & $21.4$ & $39.3$ & $50.0$ & & $16$ & $21.4 \pm 6.2$ & $37.9 \pm 8.2$ & $48.6 \pm 8.6$ \\
    & Middle Eastern & & $16$ & $31.3$ & $62.5$ & $75.0$ & & $16$ & $31.3 \pm 0.0$ & $62.5 \pm 0.0$ & $75.0 \pm 0.0$ \\
    & Southeast Asian & & $16$ & $31.3$ & $37.5$ & $56.3$ & & $16$ & $31.3 \pm 0.0$ & $37.5 \pm 0.0$ & $56.3 \pm 0.0$ \\
    & White & & $2,231$ & $30.6$ & $50.6$ & $59.5$ & & $16$ & $31.2 \pm 12.6$ & $50.8 \pm 12.3$ & $59.3 \pm 11.7$ \\

\bottomrule
\end{tabularx}
\label{tab:clip_all}
\end{table*} 

\end{document}